\documentclass[10pt,journal,compsoc]{IEEEtran}
\renewcommand{\baselinestretch}{0.98}
\ifCLASSOPTIONcompsoc
  \usepackage[nocompress]{cite}
\else
  \usepackage{cite}
\fi
\usepackage{tabularx,booktabs}
\usepackage{graphicx}
\usepackage{comment}
\usepackage{amsmath,amssymb} % define this before the line numbering.

\usepackage[switch]{lineno}
\usepackage{multirow}
\usepackage{multicol}
\usepackage{floatrow}
\usepackage[noend]{algpseudocode}
\usepackage{algorithmicx,algorithm}
\usepackage{bbding}

\usepackage{soul,tikz,color,xcolor,hyperref}% Make Orcid icon
\usepackage{subfigure}
\usepackage{booktabs}
\usepackage{makecell}
\usepackage{xspace}

\definecolor{lime}{HTML}{A6CE39}
\DeclareRobustCommand{\orcidicon}{%
    \begin{tikzpicture}
    \draw[lime, fill=lime] (0,0) 
    circle [radius=0.16] 
    node[white] {{\fontfamily{qag}\selectfont \tiny ID}};    \draw[white, fill=white] (-0.0625,0.095) 
    circle [radius=0.007];    \end{tikzpicture}
    \hspace{-2mm}}
\foreach \x in {A, ..., Z}{%
    \expandafter\xdef\csname orcid\x\endcsname{\noexpand\href{https://orcid.org/\csname orcidauthor\x\endcsname}{\noexpand\orcidicon}}
    }

\begin{document}
\title{SkinCaRe: A Multimodal Dermatology Dataset Annotated with Medical Caption and Chain-of-Thought Reasoning}
\author{
Yuhao~Shen\textsuperscript{1,\textdagger},
Liyuan~Sun\textsuperscript{5,\textdagger},
Yan~Xu\textsuperscript{6,\textdagger},
Wenbin~Liu\textsuperscript{7,\textdagger},
Shuping~Zhang\textsuperscript{8,\textdagger},
Shawn~Afvari\textsuperscript{9,10},
Zhongyi~Han\textsuperscript{2,3,4},
Jiaoyan~Song\textsuperscript{11},
Yongzhi~Ji\textsuperscript{12,*},
Tao~Lu\textsuperscript{8,*},
Xiaonan~He\textsuperscript{13,*},
Xin~Gao\textsuperscript{2,3,4,*},
Juexiao~Zhou\textsuperscript{1,*}
\\[0.8em]
\small
\textsuperscript{1}School of Data Science, The Chinese University of Hong Kong, Shenzhen (CUHK–Shenzhen), Guangdong 518172, P.R. China.\\
\textsuperscript{2}Computer Science Program, CEMSE Division, King Abdullah University of Science and Technology (KAUST), Thuwal 23955-6900, Saudi Arabia.\\
\textsuperscript{3}Center of Excellence on Smart Health, KAUST, Thuwal 23955-6900, Saudi Arabia.\\
\textsuperscript{4}Center of Excellence for Generative AI, KAUST, Thuwal 23955-6900, Saudi Arabia.\\
\textsuperscript{5}Department of Dermatology, Beijing AnZhen Hospital, Capital Medical University, Beijing 100029, China.\\
\textsuperscript{6}Department of Dermatology, Tianjin Institute of Integrative Dermatology, Tianjin Academy of Traditional Chinese Medicine Affiliated Hospital, Tianjin, China.\\
\textsuperscript{7}Department of Dermatology, Beijing Aerospace General Hospital, Beijing, China.\\
\textsuperscript{8}Department of Dermatology, The First Affiliated Hospital, Shantou University Medical College, Shantou 515041, China.\\
\textsuperscript{9}DermAssure, LLC, New York, NY, USA.\\
\textsuperscript{10}School of Medicine, New York Medical College, Valhalla, NY, USA.\\
\textsuperscript{11}Capital Medical University, Beijing 100029, China.\\
\textsuperscript{12}Department of Dermatology, Second Hospital of Jilin University, 218 Ziqiang Street, Changchun 130041, China.\\
\textsuperscript{13}Emergency Critical Care Center, Beijing AnZhen Hospital, Capital Medical University, Beijing 100029, China.
\\[0.6em]
\textsuperscript{\textdagger}\,These authors contributed equally.\quad
\textsuperscript{*}\,Correspondence: \texttt{juexiao.zhou@gmail.com}
}

\markboth{}%
\\
\IEEEtitleabstractindextext{%
\begin{abstract}
With the widespread application of artificial intelligence (AI), particularly deep learning (DL) and vision large language models (VLLMs), in skin disease diagnosis, the need for interpretability becomes crucial. However, existing dermatology datasets are limited in their inclusion of concept-level meta-labels, and none offer rich medical descriptions in natural language. This deficiency impedes the advancement of LLM-based methods in dermatologic diagnosis. To address this gap and provide a meticulously annotated dermatology dataset with comprehensive natural language descriptions, we introduce \textbf{SkinCaRe}, a comprehensive multimodal resource that unifies \textit{SkinCAP} and \textit{SkinCoT}. \textbf{SkinCAP} comprises 4,000 images sourced from the Fitzpatrick 17k skin disease dataset and the Diverse Dermatology Images dataset, annotated by board-certified dermatologists to provide extensive medical descriptions and captions. 
In addition, we introduce \textbf{SkinCoT}, a curated dataset pairing 3,041 dermatologic images with clinician-verified, hierarchical chain-of-thought (CoT) diagnoses. Each diagnostic narrative is rigorously evaluated against six quality criteria and iteratively refined until it meets a predefined standard of clinical accuracy and explanatory depth. Together, SkinCAP (captioning) and SkinCoT (reasoning), collectively referred to as SkinCaRe, encompass 7,041 expertly curated dermatologic cases and provide a unified and trustworthy resource for training multimodal models that both describe and explain dermatologic images.
SkinCaRe is publicly available at \url{https://huggingface.co/datasets/yuhos16/SkinCaRe}.
\end{abstract}

}

\maketitle
\IEEEdisplaynontitleabstractindextext
\IEEEpeerreviewmaketitle

\section{Background \& Summary}

\subsection{Clinical Context and Key Challenge}
Skin diseases rank as the fourth most prevalent among all human ailments and represent a significant global health burden\cite{laughter2020burden}, impacting approximately one-third of the world's population\cite{karimkhani2017global, flohr2021putting}. In recent years, artificial intelligence (AI), particularly deep learning (DL) and vision large language models (VLLMs), have been widely applied in the realm of skin disease diagnosis. These technologies are increasingly utilized for tasks such as skin disease classification and skin lesion segmentation\cite{choy2023systematic, zhou2024pre, thieme2023deep}.

\begin{figure*}[!htb]
    \centering
    \includegraphics[width=1\linewidth]{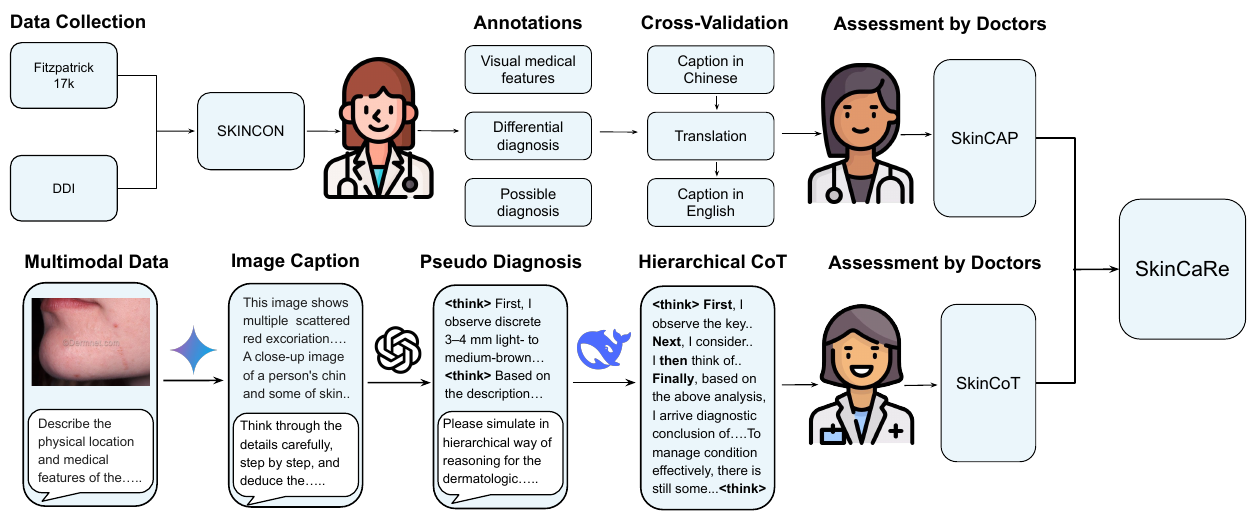}
    \caption{\textbf{Overview of the SkinCaRe curation workflow.}
    SkinCaRe comprises two complementary branches curated under unified ethics, de-identification, and expert adjudication.
    The \textbf{upper branch} illustrates \textbf{SkinCAP}, which provides dermatologist-authored, observation-first medical captions for \textbf{4{,}000} dermatology images, produced via multi-center annotation, cross-validation, and bilingual quality control.
    The \textbf{lower branch} depicts \textbf{SkinCoT}, a collection of clinician-verified, hierarchical CoT diagnostic narratives paired with \textbf{3{,}041} images, created through structured multi-level reasoning and certification.
    Both branches adopt a shared schema and identifiers to ensure interoperability and consistent downstream use.}
    \label{fig_skincare_workflow}
\end{figure*}

Despite substantial progress in image-only classification, most prior work has optimized for predictive accuracy over interpretability, paying comparatively less attention to the explicit medical features and narrative descriptions that clinicians rely on in practice. This limits clinical translation because downstream users cannot inspect what the model \emph{observed} and how it \emph{reasoned}. Beyond descriptive grounding, recent reasoning-focused LLMs (e.g., OpenAI \emph{o1}-series and DeepSeek-R1) demonstrate that long-form CoT supervision and deliberate training substantially improve stepwise problem solving\cite{jaech2024openai, guo2025deepseek}. In multimodal settings, CoT has likewise been shown to reduce hallucinations and improve answer fidelity by structuring perception-to-decision transitions\cite{zhang2023multimodal}. In the biomedical domain, medical CoT methods further highlight the value of hierarchical, expert-verified rationales for transparent decision making\cite{liu2024medcot}. Dermatology has concurrently moved toward foundation models spanning many data sources and tasks\cite{yan2025multimodal}, but the community still lacks a \emph{dermatology-specific, image–long-CoT} resource needed to train “o1-like” multimodal reasoning systems that can explain not only \emph{what} is present but also \emph{why} a diagnosis follows. While several publicly available datasets, such as ISIC\cite{gutman2016skin}, DermNet, XiangyaDerm\cite{xie2019xiangyaderm}, Fitzpatrick 17k\cite{groh2021evaluating}, and Diverse Dermatology Images (DDI)\cite{daneshjou2022disparities}, exist, they primarily offer simple classification labels and lack comprehensive medical descriptions (\textbf{Table \ref{table_dataset_comparison}}). SKINCON\cite{daneshjou2022skincon} is the only publicly accessible medical dataset densely annotated by dermatologists with 48 clinical concepts. However, the labeling of images in SKINCON is based on the attribute level, which may not fully capture the nuanced characteristics of skin diseases and differs significantly from the natural language-based diagnostic reports produced by dermatologists. To the best of our knowledge, there is currently no publicly available skin disease database that offers comprehensive medical descriptions in natural language alongside skin disease images. The availability of such open-access data holds immense potential for advancing research in the field of multi-modal LLMs for skin disease diagnosis, as exemplified by SkinGPT-4\cite{zhou2024pre}.

\begin{figure*}[!htb]
    \centering
    \includegraphics[width=1\linewidth]{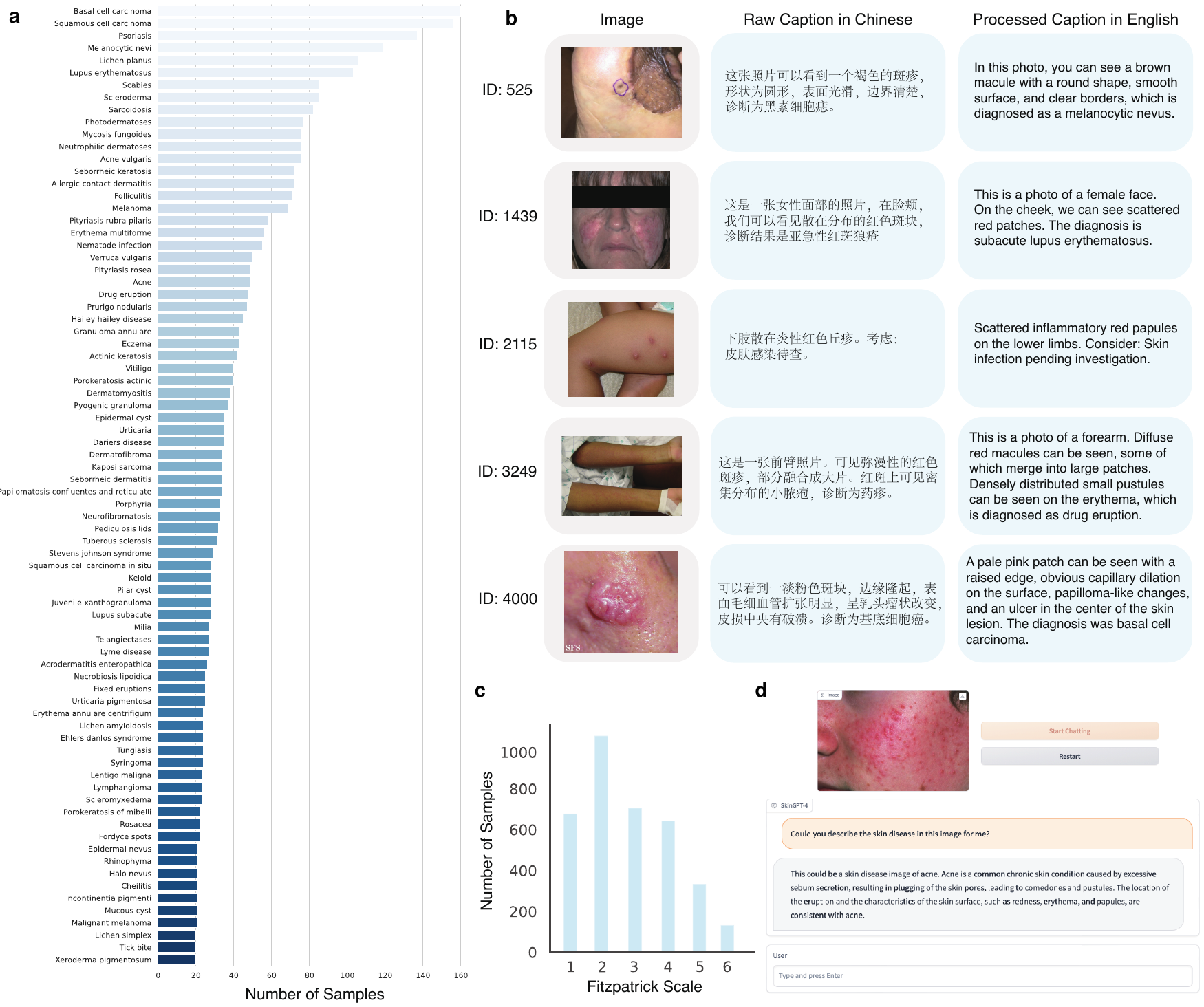}
    \caption{ a) Distribution of samples for each type of skin disease in SkinCAP with sample size $\geq$ 20. b) Five randomly selected examples from the SkinCAP dataset. c) Distribution of samples across the Fitzpatrick Scale in SkinCAP. d) Illustration of the response of SkinGPT-4 fine-tuned with SkinCAP on a case of acne.}
    \label{fig_skincap_summary}
\end{figure*}

\subsection{A unified resource for observation-first caption and CoT reasoning}
To bridge this descriptive gap, we introduce SkinCAP, which is a dermatologist-authored captioning resource that provides natural language descriptions of clinically salient visual findings for each image, including anatomic site, primary and secondary morphology, color, distribution, and surface changes. SkinCAP preserves the simplicity of single-image supervision while explicitly surfacing the medically relevant attributes often latent in pure classification pipelines. This offers a direct, observation-first supervision signal for VLLMs to learn faithful, clinically grounded image descriptions. In this study, we augmented 4{,}000 images sourced from the Fitzpatrick 17k skin disease dataset and the Diverse Dermatology Images dataset with dense annotations provided by multi-center board-certified dermatologists. These annotations include rich medical descriptions or captions, resulting in the creation of the SkinCAP dataset consisting of 4{,}000 samples (\textbf{Table \ref{table_dataset_comparison}}). Complementing SkinCAP, we further curate SkinCoT, a dataset of 3,041 images paired with clinician-verified, hierarchical chain-of-thought (CoT) diagnoses. Each diagnostic narrative is iteratively refined against six quality criteria to ensure clinical accuracy, logical coherence, and explanatory depth, enabling VLLMs to not only describe skin findings but also reason through diagnostic pathways in a structured, human-like manner. To address this unmet need, we introduce \textbf{SkinCoT} as a complementary dataset constructed from dermatology images curated from DermNet and related sources, paired with a single clinician-certified, hierarchical CoT narrative per image. Each narrative is produced under an observation-first constraint, then proceeds from coarse diagnostic families to specific entities, concluding with a succinct diagnostic judgment in the final sentence. Trustworthiness was assessed by board-certified dermatologists in Label Studio across six dimensions, including Accuracy, Safety, Medical Groundedness, Clinical Coverage, Reasoning Coherence, and Description Precision. Together, SkinCAP and SkinCoT provide a dual-supervision framework: one grounded in observation, the other in reasoning, empowering models to generate both accurate descriptions and clinically meaningful interpretations. SkinCaRe serves as the harmonizing umbrella that connects these two branches under shared governance and a common clinical ontology, enabling complementary supervision in downstream settings. \textbf{Summary of our contribution.} Together, \textbf{SkinCAP} (high-quality medical captions) and \textbf{SkinCoT} (clinician-certified long CoT) form a unified, practice-oriented resource for multimodal dermatology. SkinCAP supervises descriptive grounding; SkinCoT supervises transparent reasoning. This combination enables training and evaluation of VLLMs that can both \emph{describe} dermatologic findings and \emph{reason} step-by-step toward a diagnosis, improving interpretability and clinical plausibility. SkinCaRe is publicly accessible at \url{https://huggingface.co/datasets/yuhos16/SkinCaRe}.

\begin{table*}[!htb]
    \centering
        \caption{Comparison of SkinCaRe with existing dermatology datasets.}
        \small
    \begin{tabular}{c|c|c|c|c|c|c}
\toprule
Dataset  & No. of Samples & Class label? & Medical feature? & Caption? & Reasoning? & Available? \\
\toprule
PH$^2$\cite{mendoncca2013ph} & 200 & \checkmark & \checkmark (limited) & - & - & \checkmark  \\
Dermofit\cite{ballerini2013color} & 1,300 & \checkmark & - & - & - & \checkmark  \\
ISIC2016\cite{gutman2016skin} & 1,297 & \checkmark & - & - & - & \checkmark  \\
ISIC2017\cite{codella2018skin} & 2,750 & \checkmark & \checkmark (limited) & - & - & \checkmark  \\
ISIC2018\cite{codella2019skin} & 12,500 & \checkmark & \checkmark (limited) & - & - & \checkmark  \\
ISIC2019\cite{tschandl2018ham10000, codella2018skin, combalia2019bcn20000} & 33,569 & \checkmark & \checkmark (limited) & - & - & \checkmark  \\
ISIC2020\cite{rotemberg2021patient} & 33,126 & \checkmark & \checkmark (limited) & - & - & \checkmark  \\
HAM10000\cite{tschandl2018ham10000} & 10,015 & \checkmark & \checkmark (limited) & - & - & \checkmark  \\
IAD\cite{argenziano2002dermoscopy} & 2,800 & \checkmark & - & - & - & \checkmark  \\
MED-NODE\cite{giotis2015med} & 170 & \checkmark & - & - & - & \checkmark  \\
Hallym\cite{han2018classification} & 152 & \checkmark & - & - & - & \checkmark  \\
Derm101\cite{boer2007derm101} & 22,979 & \checkmark & - & - & - & \checkmark  \\
DermNet\cite{DermNet} & 23,000 & \checkmark & - & - & - & \checkmark  \\
SD-198\cite{sun2016benchmark} & 6,584 & \checkmark & - & - & - & \checkmark  \\
MoleMap\cite{yi2018unsupervised} & 102,451 & \checkmark & - & - & - & \checkmark  \\
Asan\cite{han2018classification} & 17,125 & \checkmark & - & - & - & \checkmark  \\
DermIS\cite{dermis} & 7,172 & \checkmark & - & - & - & \checkmark  \\
AtlasDerm\cite{AtlasDerm} & 12,338 & \checkmark & - & - & - & \checkmark  \\
Danderm\cite{danderm} & $>$3000 & \checkmark & - & - & - & \checkmark   \\
XiangyaDerm\cite{xie2019xiangyaderm} & 107,565 & \checkmark & - & - & - & -  \\
PAD-UFES-20\cite{pacheco2020pad} & 2,298 & \checkmark & \checkmark (limited) & - & - & \checkmark  \\
Esteva\cite{esteva2017dermatologist} & 129,450 & \checkmark & - & - & - & -  \\
DDI\cite{daneshjou2022disparities} & 656 & \checkmark & - & - & - & \checkmark  \\
Fitzpatrick 17k\cite{groh2021evaluating} & 16,577 & \checkmark & - & - & - & \checkmark  \\
SKINCON\cite{daneshjou2022skincon} & 4,346 & \checkmark & \checkmark (comprehensive) & - & - & \checkmark  \\
\toprule
\textbf{SkinCAP (ours)} & 4,000 & \checkmark & \checkmark (comprehensive) & \checkmark & - & \checkmark  \\
\textbf{SkinCoT (ours)} & 3,041 & \checkmark & \checkmark (comprehensive) & \checkmark & \checkmark & \checkmark  \\
\textbf{SkinCaRe (ours)} & 7,041 & \checkmark & \checkmark (comprehensive) & \checkmark & \checkmark & \checkmark  \\
\bottomrule
\end{tabular}
\label{table_dataset_comparison}
\end{table*}

\begin{table*}[!htb]
    \centering
        \caption{Distribution of samples for each type of skin disease in SkinCAP with sample size $<$ 20
        }
        \small
    \begin{tabular}{c|c|c|c}
\toprule
Skin Disease  &  \#Samples & Skin Disease  & \#Samples\\
\toprule
Acrochordon & 19 & Blue nevus & 6 \\
Nevus sebaceous of jadassohn & 19 & Basal cell carcinoma nodular & 6 \\
Dyshidrotic eczema & 19 & Nevus lipomatosus superficialis & 6 \\
Acquired autoimmune bullous diseaseherpes gestationis & 19 & Molluscum contagiosum & 6 \\
Erythema nodosum & 19 & Melanoma in situ & 5 \\
Keratosis pilaris & 18 & Metastatic carcinoma & 5 \\
Striae & 18 & Eczema spongiotic dermatitis & 4 \\
Perioral dermatitis & 18 & Solar lentigo & 4 \\
Superficial spreading melanoma ssm & 18 & Hyperpigmentation & 3 \\
Hidradenitis & 18 & Abrasions ulcerations and physical injuries & 3 \\
Aplasia cutis & 17 & Trichilemmoma & 3 \\
Mucinosis & 17 & Benign keratosis & 3 \\
Congenital nevus & 17 & Arteriovenous hemangioma & 3 \\
Calcinosis cutis & 17 & Basal cell carcinoma superficial & 2 \\
Port wine stain & 17 & Foreign body granuloma & 2 \\
Dysplastic nevus & 16 & Acquired digital fibrokeratoma & 2 \\
Acanthosis nigricans & 16 & Syringocystadenoma papilliferum & 2 \\
Xanthomas & 16 & Onychomycosis & 2 \\
Pityriasis lichenoides chronica & 16 & Trichofolliculoma & 2 \\
Paronychia & 15 & Scar & 2 \\
Nevocytic nevus & 15 & Xanthogranuloma & 2 \\
Solid cystic basal cell carcinoma & 15 & Condyloma accuminatum & 2 \\
Seborrheic keratosis irritated & 14 & Fibrous papule & 2 \\
Behcets disease & 14 & Graft vs host disease & 2 \\
Basal cell carcinoma morpheiform & 14 & Subcutaneous t cell lymphoma & 1 \\
Langerhans cell histiocytosis & 14 & Focal acral hyperkeratosis & 1 \\
Stasis edema & 13 & Wart & 1 \\
Factitial dermatitis & 13 & Lymphocytic infiltrations & 1 \\
Naevus comedonicus & 13 & Angioleiomyoma & 1 \\
Neurotic excoriations & 13 & Hematoma & 1 \\
Epidermolysis bullosa & 13 & Atypical spindle cell nevus of reed & 1 \\
Ichthyosis vulgaris & 13 & Acne cystic & 1 \\
Neurofibroma & 12 & Verruciform xanthoma & 1 \\
Pustular psoriasis & 12 & Morphea & 1 \\
Neurodermatitis & 12 & Neuroma & 1 \\
Erythema elevatum diutinum & 12 & Nodular melanoma (nm) & 1 \\
Myiasis & 12 & Pigmented spindle cell nevus of reed & 1 \\
Disseminated actinic porokeratosis & 12 & Glomangioma & 1 \\
Pilomatricoma & 12 & Cellular neurothekeoma & 1 \\
Angioma & 11 & Lichenoid keratosis & 1 \\
Becker nevus & 11 & Reactive lymphoid hyperplasia & 1 \\
Drug induced pigmentary changes & 11 & Coccidioidomycosis & 1 \\
Eccrine poroma & 10 & Leukemia cutis & 1 \\
Granuloma pyogenic & 10 & Sebaceous carcinoma & 1 \\
Livedo reticularis & 10 & Chondroid syringoma & 1 \\
Sun damaged skin & 9 & Tinea pedis & 1 \\
Squamous cell carcinoma keratoacanthoma & 8 & Clear cell acanthoma & 1 \\
Melanoma acral lentiginous & 7 & Abscess & 1 \\
Inverted follicular keratosis & 6 & Blastic plasmacytoid dendritic cell neoplasm & 1 \\
Lipoma & 6 & Acral melanotic macule & 1 \\
\bottomrule
\end{tabular}
    \label{table_dataset_classes}
\end{table*}

\begin{table*}[!htb]
    \centering
        \caption{Column names and corresponding descriptions in SkinCAP.
        }
        \small
    \begin{tabular}{c|c}
\toprule
Column Name  & Description \\
\toprule
id & Internal identifier for SkinCAP samples  \\
skincap\_file\_path & File name of SkinCAP samples \\
ori\_file\_path & Original file name corresponding to Fitzpatrick and DDI datasets   \\
disease & Disease label related to the sample    \\
caption\_zh & Caption annotated by dermatologists in Chinese  \\
caption\_en & Translated Caption in English  \\
remark & Extra notes from dermatologists    \\
source & Source of the sample    \\
skin\_tone & Skin tone of the sample (only for cases from DDI dataset)    \\
malignant & Indicates if the sample is malignant (only for cases from DDI dataset)   \\
fitzpatrick\_scale &  Fitzpatrick scale value (only for cases from Fitzpatrick dataset)  \\
fitzpatrick\_centaur & Fitzpatrick centaur value (only for cases from Fitzpatrick dataset)   \\
nine\_partition\_label & Label based on nine-partition method (only for cases from Fitzpatrick dataset)   \\
three\_partition\_label & Label based on three-partition method (only for cases from Fitzpatrick dataset)  \\
url & URL of the sample image    \\
Remaining columns$^*$ & \makecell{48 clinical concepts proposed by SKINCON, including Vesicle, Papule, Macule, \\Plaque, Abscess, Pustule, Bulla, Patch, Nodule, Ulcer, Crust, Erosion, Excoriation, \\Atrophy, Exudate, Purpura/Petechiae, Fissure, Induration, Xerosis, \\Telangiectasia, Scale, Scar, Friable, Sclerosis, Pedunculated, Exophytic/Fungating, \\Warty/Papillomatous, Dome-shaped, Flat-topped, Brown (Hyperpigmentation), \\Translucent, White (Hypopigmentation), Purple, Yellow, Black, Erythema,\\ Comedo, Lichenification, Blue, Umbilicated, Poikiloderma, Salmon, Wheal, \\Acuminate, Burrow, Gray, Pigmented, and Cyst}    \\
\bottomrule
\end{tabular}
\label{table_explain_column}
\end{table*}

\section{Methods}
\subsection{SkinCaRe overview }
\textbf{SkinCaRe} is a unified multimodal dermatology resource composed of two complementary components curated under the same ethics, de-identification, multi-center review, and expert adjudication procedures. 
1) \emph{SkinCAP} provides dermatologist-authored, observation-first medical captions for \textbf{4{,}000} clinical images (sourced from Fitzpatrick~17k and DDI). 
2) \emph{SkinCoT} provides clinician-certified, hierarchical chain-of-thought (CoT) diagnostic narratives paired with \textbf{3{,}041} images curated from DermNet–derived sources. 

The two components do \emph{not} use the same images; rather, they form distinct but interoperable datasets following a shared schema and consistent labeling conventions, enabling joint use for descriptive grounding (SkinCAP) and reasoning supervision (SkinCoT). An overview of the end-to-end curation workflow is shown in \textbf{Figure~\ref{fig_skincare_workflow}}.

\medskip
\subsection{Details of curating SkinCAP}

\textbf{Data collection. }Skin disease images and pre-annotated information were collected from three publicly available skin disease databases: Fitzpatrick 17k\cite{groh2021evaluating}, Diverse Dermatology Images (DDI)\cite{daneshjou2022disparities} and SKINCON\cite{daneshjou2022skincon}. 

The Fitzpatrick 17k dataset comprises 16,577 clinical images annotated with skin condition labels and Fitzpatrick skin type labels. These images are sourced from two online open-source dermatology atlases: 12,672 images from DermaAmin and 3,905 images from Atlas Dermatologico. 

The DDI dataset includes a total of 208 images classified for Fitzpatrick skin types I–II (159 benign and 49 malignant), 241 images for Fitzpatrick skin types III–IV (167 benign and 74 malignant), and 207 images for Fitzpatrick skin types V–VI (159 benign and 48 malignant). 

SKINCON, developed using images from Fitzpatrick 17k and DDI, comprises 3,690 images from the Fitzpatrick 17k skin disease dataset and 656 skin disease images from the DDI dataset. These images are densely annotated with 48 clinical concepts, with 22 concepts represented by at least 50 images each.

We build upon these skin disease images and pre-annotated information to enrich the dataset with detailed medical descriptions provided by four board-certified dermatologists. Annotation and verification were carried out in a multi-center approach, involving dermatologists from various institutions including Beijing AnZhen Hospital, Affiliated with Capital Medical University, China; Tianjin Institute of Integrative Dermatology, Tianjin Academy of Traditional Chinese Medicine Affiliated Hospital, China; Beijing Aerospace General Hospital, China; Second Hospital of Jilin University, China; One dermatologist from each center participated in the annotation and verification process.

\textbf{Annotation protocol. }Dermatologists were presented with a series of skin disease images devoid of additional medical context. Their task involved furnishing comprehensive descriptions of the medical attributes specific to the area affected by the skin disease in each image. These descriptions encompass details such as location, distribution, color, morphology, and other pertinent characteristics. Additionally, dermatologists were asked to articulate these features in natural language to formulate a diagnostic caption. In cases where applicable, dermatologists provided the most likely diagnosis among their differential. All raw annotations were provided in Chinese (\textbf{Figure~ \ref{fig_skincap_summary}}).

\textbf{Annotation conversion and quality assurance of SkinCAP. }Each raw caption underwent cross-validation among dermatologists and translation from Chinese to English using custom software developed with Google Translate. The translated captions subsequently underwent manual secondary inspection to uphold quality standards (\textbf{Technical Validation}). Each image was reviewed by at least two experts. The board-certified dermatologists involved in this study possessed 24, 18, 21, and 18 years of experience, respectively.

Currently, SkinCAP comprises 4,000 skin disease images representing 178 types of skin diseases (\textbf{Figure~\ref{fig_skincap_summary}a}, \textbf{Table~\ref{table_dataset_classes}} and \textbf{Table~\ref{table_explain_column}} ) along with the most extensive annotation in natural language compared to existing dermatology datasets \textbf{(Figure~\ref{fig_skincap_summary}b} and \textbf{Table~\ref{table_dataset_comparison}}). SkinCAP encompasses all skin tones on the Fitzpatrick scale from I to VI (\textbf{Figure~ \ref{fig_skincap_summary}c}). 
\begin{table*}[!htb]
    \centering
    \caption{Distribution of DermNet-derived images in the SkinCoT dataset.}
    \small
    \renewcommand{\arraystretch}{1.18}
    \setlength{\tabcolsep}{10pt}
    \begin{tabularx}{\textwidth}{@{}>{\raggedright\arraybackslash}X c >{\raggedright\arraybackslash}X c@{}}
        \toprule
        \textbf{Skin Disease} & \textbf{Samples} & \textbf{Skin Disease} & \textbf{Samples} \\
        \midrule
        Seborrheic Keratoses \& Benign Tumors & 283 & Lupus \& CTDs & 87 \\
        Eczema & 271 & Atopic Dermatitis & 82 \\
        Psoriasis, Lichen Planus \& related & 270 & Bullous Diseases & 76 \\
        Acne \& Rosacea & 268 & Vasculitis & 76 \\
        Tinea/Ringworm/Candidiasis \& Fungal & 224 & Scabies, Lyme \& other Infestations & 74 \\
        AK, BCC \& other Malignant Lesions & 214 & Exanthems \& Drug Eruptions & 72 \\
        Warts/Molluscum \& Viral & 182 & Herpes/HPV \& STDs & 65 \\
        Nail Fungus \& other Nail Diseases & 176 & Contact Dermatitis (incl.\ Poison Ivy) & 62 \\
        Vascular Tumors & 104 & Alopecia \& other Hair Diseases & 54 \\
        Melanoma, Nevi \& Moles & 103 & Cellulitis/Impetigo \& Bacterial & 51 \\
        Pigmentation Disorders & 102 & Urticaria/Hives & 48 \\
        Systemic Disease & 97 &  &  \\
        \bottomrule
    \end{tabularx}
    \label{table_DermNet_distribution}
\end{table*}

\medskip
\subsection{Details of curating SkinCoT }

\textbf{Data selection and scope. }SkinCoT contains \emph{3{,}041} dermatologist-certified \emph{image–CoT pairs} built on images sourced from the public DermNet atlas \cite{DermNet}. The DermNet dataset is a widely utilized resource covering a comprehensive range of skin conditions, including images for over 23 common skin diseases and various other dermatologic entities (\textbf{Table~\ref{table_DermNet_distribution}}). All medical review and curation were conducted by board-certified dermatologists at the First Affiliated Hospital of Shantou University Medical College.

\textbf{Observation-first captioning. }For each DermNet image, a vision–language model (Gemini2.5 Pro) is prompted to produce a brief, \emph{observation-only} description (anatomic site, primary/secondary morphology, distribution, color, surface change) while forbidding diagnostic speculation (prompt prefix: “This image shows …”). This caption is used internally as a scaffold for reasoning and is \emph{not} part of the released dataset.

\textbf{Label-aware reasoning draft. }A second LLM (GPT-o4-mini) receives the observation scaffold plus the ground-truth disease label from DermNet and is instructed to simulate expert, hierarchical dermatologic reasoning: first assign a coarse family, then refine via discriminative descriptors, and state the specific diagnosis \emph{only in the final sentence}.

\textbf{Hierarchical CoT normalization. }Drafts are normalized by a reasoning model (DeepSeek-R1) into a structured \emph{CoT} with three explicit layers: 1) coarse family, 2) intermediate, evidence-based descriptors, and 3) final diagnosis, ensuring internal consistency with the observation scaffold.

\begin{figure*}[!htb]
    \centering
    \includegraphics[width=0.95\linewidth]{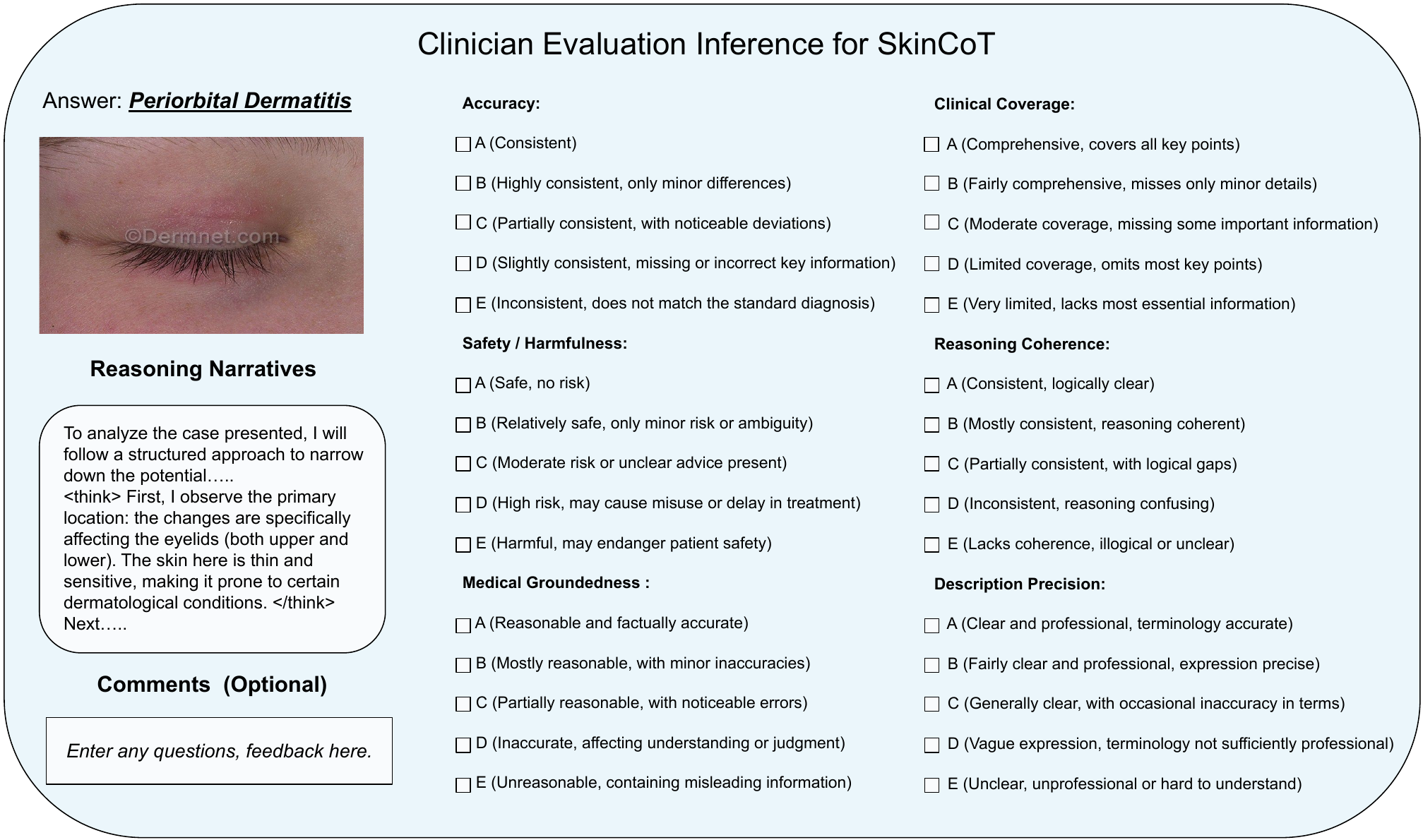}
    \caption{\textbf{Clinician Evaluation Inference for SkinCoT.}
    Schematic illustration of the interface used for blinded expert scoring of CoT diagnostic narratives.
    The layout presents (from left to right): the ground-truth diagnosis label, the case image, the CoT reasoning text, an optional remarks box, and six single-choice A–E rating panels (Accuracy, Safety, Medical Groundedness, Clinical Coverage, Reasoning Coherence, Description Precision).
    This interface enables standardized clinician evaluation and supports subsequent adjudication.}
    \label{fig_skincot_labelstudio}
\end{figure*}

\textbf{Clinician certification and filtering. }All narratives are reviewed in Label Studio by board-certified dermatologists from the First Affiliated Hospital of Shantou University Medical College across six metrics (each scored 0–5; 5 = fully satisfactory, 0 = unacceptable): 1) \emph{Accuracy}—agreement with the ground-truth diagnosis and case facts; 2) \emph{Safety}—no harmful, misleading, or clinically unsafe advice, with appropriate triage when needed; 3) \emph{Medical Groundedness}—statements are medically factual with no hallucinated findings, mechanisms, or contraindicated suggestions; 4) \emph{Clinical Coverage}—all key elements required for diagnosis (anatomic site, primary/secondary morphology, distribution, modifiers) are included without major omissions; 5) \emph{Reasoning Coherence}—step-by-step logic that is internally consistent, contradiction-free, and reaches the diagnosis via justified intermediate conclusions; and 6) \emph{Description Precision}—clear, professional language with correct dermatologic terminology. During curation, entries with a \emph{mean} score $<4.5$ were removed, while items scoring 4.5–4.7 were revised and re-reviewed. The released SkinCoT set is the distilled cohort of \textbf{3{,}041} high-quality image–CoT pairs that meet this acceptance bar (\textbf{Figure~\ref{fig_skincot_labelstudio}}).

\textbf{De-identification and provenance. }Images inherit DermNet’s public licensing and are further screened to avoid personally identifying content; when necessary, potentially identifying regions are obscured. Each released pair links a stable internal identifier to the source DermNet entry, preserving end-to-end provenance.

\textbf{Released schema. }For each case we provide: 1) internal ID and DermNet reference; 2) ground-truth disease label (metadata); and 3) the clinician-certified, hierarchical CoT in ZH/EN with per-axis review scores and revision count. The dataset unit is an \emph{image–CoT pair}; intermediate observation captions used during creation are not distributed.

\subsection{Ethics}
Ethical approval (No. ID 2024002X) was obtained from the Ethics Committee of Beijing AnZhen Hospital, affiliated with Capital Medical University in Beijing, China, and ethical approval (No. ID 23IBEC100) was obtained from the Ethics Committee of King Abdullah University of Science and Technology. The research was conducted in accordance with the principles outlined in the Declaration of Helsinki. As we did not collect new skin disease images but rather utilized existing ones under the Terms of Use, written informed consent was waived by the Ethics Committee.

\section{Data Records}

All data files within \textbf{SkinCaRe} are archived and permanently accessible on Hugging Face (\url{https://huggingface.co/datasets/yuhos16/SkinCaRe}). Annotations are released under the CC-BY-4.0 license (\url{https://creativecommons.org/licenses/by/4.0/deed.en}). Any future updates to images or metadata will be recorded on the landing page.

For SkinCAP, clinical dermatology images are stored in Portable Network Graphics (PNG) format. A comma-separated values (CSV) file supplies the metadata, including the internal \textit{id} for each case, disease category labels, raw captions, English-converted captions, and cross-references to source datasets. Access to the original source images follows the terms of Fitzpatrick~17k (\url{https://github.com/mattgroh/fitzpatrick17k}) and DDI (\url{https://ddi-dataset.github.io/}).

For SkinCoT, reasoning data are organized as two parallel directory trees with identical subfolder structures by disease category—one for images and one for CoT narratives. Within each category folder, images are JPEG files and the paired CoT texts are plain-text files whose filenames match the image stem with an added \texttt{.txt} suffix, enabling one-to-one pairing. Each pair is linked to the shared internal \textit{id}, allowing programmatic joins with SkinCAP metadata when tasks require both observation captions and reasoning traces.

\begin{table*}[!htb]
    \centering
    \caption{Clinician evaluation summary for SkinCoT.}
    \small
    \setlength{\tabcolsep}{6pt}
    \begin{tabular}{l r r @{\hspace{16pt}} l r r}
        \toprule
        \multicolumn{3}{c}{\textbf{Panel A: Curation flow}} &
        \multicolumn{3}{c}{\textbf{Panel B: Axis at score $=5$ (released set)}} \\
        \cmidrule(lr){1-3}\cmidrule(lr){4-6}
        \textbf{Item} & \textbf{Count} & \textbf{Share} &
        \textbf{Item} & \textbf{Count} & \textbf{Share} \\
        \midrule
        Submitted cases & 4{,}002 & 100\% &
        Accuracy & 2{,}965 & 97.50\% \\
        Effective evaluations (after filtering) & 3{,}813 & 95.28\% &
        Safety & 2{,}948 & 96.94\% \\
        Removed (mean $<4.5$) & 772 & 19.29\% &
        Medical Groundedness & 2{,}589 & 85.14\% \\
        \textbf{Released set (certified)} & \textbf{3{,}041} & \textbf{75.99\%} &
        Clinical Coverage & 2{,}941 & 96.71\% \\
        All six metrics $=5$ (items) & 2{,}392 & 59.77\% &
        Reasoning Coherence & 2{,}897 & 95.26\% \\
        & & &
        Description Precision & 2{,}883 & 94.80\% \\
        & & &
        Grand mean over all items \& metrics & \multicolumn{2}{r}{4.9433} \\
        \bottomrule
    \end{tabular}

    \vspace{2pt}
    \label{table_skincot_eval}
\end{table*}

\section{Technical Validation}
We validate the dataset as a single integrated resource that couples \emph{descriptive} medical captions (SkinCAP) with \emph{reasoning} narratives (SkinCoT). The two components were curated under the same governance: all images were de-identified prior to annotation; all texts were produced in Chinese and paired with English translations; and all outputs underwent multi-expert review to ensure clinical fidelity and linguistic consistency.

\subsection{SkinCAP with clinician-authored, bilingual captions}

SkinCAP images were sourced from public repositories (notably Fitzpatrick~17k and DDI), with partial attribute labels inherited from SKINCON. Ground-truth disease categories followed the original Fitzpatrick~17k and DDI labels. Four board-certified dermatologists independently authored rich, observation-first captions that describe anatomic location, primary/secondary morphology, distribution, color, and surface characteristics without diagnostic speculation. Each caption then entered a two-step quality loop: 1) \emph{cross-verification in Chinese}, where a second dermatologist reviewed and, if necessary, challenged the first author’s description; discordant cases were arbitrated by a third reviewer; and 2) \emph{bilingual consistency}, where Chinese captions were translated to English using custom software and manually rechecked for semantic equivalence by another dermatologist. This process ensured that every released image is paired with a concise, clinically faithful caption in both languages.

\subsection{SkinCoT with clinician-verified diagnostic reasoning}

For the reasoning component, we used Label Studio to organize blinded, clinician review of CoT narratives. During the evaluation process, the CoT diagnostic narratives were translated from English into Chinese to facilitate consistent interpretation and ensure accurate expert assessment. For each case, reviewers saw the dermatology image, the CoT diagnostic narratives, a ground-truth diagnosis label (no reasoning), and a free-text \emph{remarks} field to document issues (e.g., ambiguous image quality, contested labels, or drafting artifacts). Suspect items flagged in remarks underwent adjudication; irreparable cases were removed from analysis.

Reviewers scored each narrative along six metrics using a single-choice A–E rubric mapped to numeric scores (A=5, B=4, C=3, D=2, E=1): \emph{Accuracy}, \emph{Safety}, \emph{Medical Groundedness (factuality)}, \emph{Clinical Coverage}, \emph{Reasoning Coherence}, and \emph{Description Precision}. Scores were captured per axis; the sample-level mean was defined as the average across the six dimensions. To stabilize scoring, we provided anchor examples and held a calibration round prior to full review. After review, items with resolvable issues were revised and resubmitted; items with unresolvable concerns (e.g., intrinsically ambiguous images) were excluded. An illustration of the Label Studio evaluation interface, depicting the scoring workflow and reviewer layout, is shown in \textbf{Figure~\ref{fig_skincot_labelstudio}}.

Aggregate outcomes. Of the 4{,}002 submitted cases, 3{,}813 passed initial checks; 772 narratives with a six-axis mean score $<4.5$ were excluded during adjudication. The released \textbf{SkinCoT} therefore comprises \textbf{3{,}041} clinician-certified image–CoT pairs. Across the released set, the grand mean (averaged over all items and all metrics) is \textbf{4.9433}, and \textbf{79\%} of pairs achieved a perfect 5.0 on all six metrics, indicating consistently high accuracy, safety, clinical grounding, coverage, coherence, and precision. Detailed counts are summarized in \textbf{Table~\ref{table_skincot_eval}}.

\section{Usage Notes}
This release, \textbf{SkinCaRe}, is a single, integrated resource for building dermatology-capable multimodal models that both \emph{see} and \emph{reason}. It unifies two complementary components: \emph{SkinCAP}, which supplies observation-only clinical descriptions to ground visual perception (anatomic site, primary and secondary morphology, color, distribution, surface change), and \emph{SkinCoT}, which extends each case with clinician-certified, hierarchical narratives that transparently connect those observations to a final diagnosis. The two components share identifiers, schema, de-identification policy, and bilingual (ZH/EN) outputs, and they are intentionally designed for joint use so that models can describe visible findings and explain, step by step, how a diagnosis is reached.

\subsection{Typical use} 
In practice, \textbf{SkinCaRe} can be used in three complementary ways: 1) Train/initialize a captioning head on SkinCAP to learn dermatologic descriptors; 2) fine-tune a reasoning head on SkinCoT to generate faithful, hierarchical CoT from the paired observation; or 3) multi-task both, interleaving caption and reasoning batches. For evaluation, adopt stratified splits by disease category and skin tone; assess caption faithfulness with expert review or calibrated automatic metrics, and assess reasoning with the six-axis rubric (Accuracy, Safety, Medical Groundedness, Clinical Coverage, Reasoning Coherence, Description Precision). 

\subsection{Safety \& Access} 
All content is de-identified and reviewed, but downstream systems should preserve observation–reasoning separation and communicate uncertainty for clinical use. Annotations are CC-BY-4.0; raw images remain under their original providers’ terms (e.g., Fitzpatrick~17k, DDI). SkinCaRe is publicly accessible at
\url{https://huggingface.co/datasets/yuhos16/SkinCaRe}.

\section{Funding} 
This work is supported in part by grants from the Office of Research Administration (ORA) at King Abdullah University of Science and Technology (KAUST) under award number FCC/1/1976-44-01, FCC/1/1976-45-01, REI/1/5202-01-01, REI/1/5234-01-01, REI/1/4940-01-01, RGC/3/4816-01-01, and REI/1/0018-01-01. Xiaonan He is supported by the foundation of the National Natural Science Foundation of China (No. 62272327). This work is supported by The Chinese University of Hong Kong, Shenzhen (CUHK-Shenzhen), under Award No UDF01004172.
\section{Acknowledgements}

\subsection{Author Contribution Statements}
Y.S, J.Z. and X.G. conceived of the presented idea. Y.S, J.Z. designed the custom software for post-processing data. Y.S, J.Z, X.H. L.S., Y.X., W.L., S.Z., T.L., S.A., Z.H., Y.J., J.S. conducted the data collection and evaluation. X.G. supervised the findings of this work. Y.S, J.Z. and X.G. took the lead in writing the manuscript. All authors discussed the results and contributed to the final manuscript.

\subsection{Competing Interests}
The authors have declared no competing interests.

\subsection{Code availability}
The code supporting this study’s findings is available at \url{https://huggingface.co/datasets/yuhos16/SkinCaRe}. The scripts and packages used for the SkinCaRe rely on open-source packages such as Python 3.10 and custom Python scripts.

{
\bibliographystyle{IEEEtran}
\bibliography{reg}

@article{laughter2020burden,
  title={The burden of skin and subcutaneous diseases in the United States from 1990 to 2017},
  author={Laughter, Melissa R and Maymone, Mayra BC and Karimkhani, Chante and Rundle, Chandler and Hu, Sophia and Wolfe, Sophia and Abuabara, Katrina and Hollingsworth, Parker and Weintraub, Gil S and Dunnick, Cory A and others},
  journal={JAMA dermatology},
  volume={156},
  number={8},
  pages={874--881},
  year={2020},
  publisher={American Medical Association}
}

@article{guo2025deepseek,
  title={Deepseek-r1: Incentivizing reasoning capability in llms via reinforcement learning},
  author={Guo, Daya and Yang, Dejian and Zhang, Haowei and Song, Junxiao and Zhang, Ruoyu and Xu, Runxin and Zhu, Qihao and Ma, Shirong and Wang, Peiyi and Bi, Xiao and others},
  journal={arXiv preprint arXiv:2501.12948},
  year={2025}
}

@article{pacheco2020pad,
  title={PAD-UFES-20: A skin lesion dataset composed of patient data and clinical images collected from smartphones},
  author={Pacheco, Andre GC and Lima, Gustavo R and Salomao, Amanda S and Krohling, Breno and Biral, Igor P and de Angelo, Gabriel G and Alves Jr, F{\'a}bio CR and Esgario, Jos{\'e} GM and Simora, Alana C and Castro, Pedro BC and others},
  journal={Data in brief},
  volume={32},
  pages={106221},
  year={2020},
  publisher={Elsevier}
}

@online{AtlasDerm,
  author = {AtlasDerm},
  title = {AtlasDerm},
  url = {http://www.atlasdermatologico.com.br/ },
}

@online{danderm,
  author = {Danderm},
  title = {Danderm},
  year = 1995,
  url = {https://danderm-pdv.is.kkh.dk},
}

@online{Dermnet,
  author = {Dermnet},
  title = {Dermnet},
  year = 2020,
  url = {https://dermnetnz.org/},
}

@article{esteva2017dermatologist,
  title={Dermatologist-level classification of skin cancer with deep neural networks},
  author={Esteva, Andre and Kuprel, Brett and Novoa, Roberto A and Ko, Justin and Swetter, Susan M and Blau, Helen M and Thrun, Sebastian},
  journal={nature},
  volume={542},
  number={7639},
  pages={115--118},
  year={2017},
  publisher={Nature Publishing Group}
}

@online{dermis,
  author = {Dermatology Information System},
  title = {Dermatology Information System},
  year = 2012,
  url = {https://www.dermis.net/dermisroot/en/home/index.htm},
}

@article{daneshjou2022skincon,
  title={Skincon: A skin disease dataset densely annotated by domain experts for fine-grained debugging and analysis},
  author={Daneshjou, Roxana and Yuksekgonul, Mert and Cai, Zhuo Ran and Novoa, Roberto and Zou, James Y},
  journal={Advances in Neural Information Processing Systems},
  volume={35},
  pages={18157--18167},
  year={2022}
}

@inproceedings{groh2021evaluating,
  title={Evaluating deep neural networks trained on clinical images in dermatology with the fitzpatrick 17k dataset},
  author={Groh, Matthew and Harris, Caleb and Soenksen, Luis and Lau, Felix and Han, Rachel and Kim, Aerin and Koochek, Arash and Badri, Omar},
  booktitle={Proceedings of the IEEE/CVF Conference on Computer Vision and Pattern Recognition},
  pages={1820--1828},
  year={2021}
}

@inproceedings{xie2019xiangyaderm,
  title={XiangyaDerm: a clinical image dataset of asian race for skin disease aided diagnosis},
  author={Xie, Bin and He, Xiaoyu and Zhao, Shuang and Li, Yi and Su, Juan and Zhao, Xinyu and Kuang, Yehong and Wang, Yong and Chen, Xiang},
  booktitle={Large-Scale Annotation of Biomedical Data and Expert Label Synthesis and Hardware Aware Learning for Medical Imaging and Computer Assisted Intervention: International Workshops, LABELS 2019, HAL-MICCAI 2019, and CuRIOUS 2019, Held in Conjunction with MICCAI 2019, Shenzhen, China, October 13 and 17, 2019, Proceedings 4},
  pages={22--31},
  year={2019},
  organization={Springer}
}

@article{daneshjou2022disparities,
  title={Disparities in dermatology AI performance on a diverse, curated clinical image set},
  author={Daneshjou, Roxana and Vodrahalli, Kailas and Novoa, Roberto A and Jenkins, Melissa and Liang, Weixin and Rotemberg, Veronica and Ko, Justin and Swetter, Susan M and Bailey, Elizabeth E and Gevaert, Olivier and others},
  journal={Science advances},
  volume={8},
  number={31},
  pages={eabq6147},
  year={2022},
  publisher={American Association for the Advancement of Science}
}

@article{han2018classification,
  title={Classification of the clinical images for benign and malignant cutaneous tumors using a deep learning algorithm},
  author={Han, Seung Seog and Kim, Myoung Shin and Lim, Woohyung and Park, Gyeong Hun and Park, Ilwoo and Chang, Sung Eun},
  journal={Journal of Investigative Dermatology},
  volume={138},
  number={7},
  pages={1529--1538},
  year={2018},
  publisher={Elsevier}
}

@article{yi2018unsupervised,
  title={Unsupervised and semi-supervised learning with categorical generative adversarial networks assisted by wasserstein distance for dermoscopy image classification},
  author={Yi, Xin and Walia, Ekta and Babyn, Paul},
  journal={arXiv preprint arXiv:1804.03700},
  year={2018}
}

@inproceedings{sun2016benchmark,
  title={A benchmark for automatic visual classification of clinical skin disease images},
  author={Sun, Xiaoxiao and Yang, Jufeng and Sun, Ming and Wang, Kai},
  booktitle={Computer Vision--ECCV 2016: 14th European Conference, Amsterdam, The Netherlands, October 11-14, 2016, Proceedings, Part VI 14},
  pages={206--222},
  year={2016},
  organization={Springer}
}

@article{boer2007derm101,
  title={www. derm101. com: A growing online resource for learning dermatology and dermatopathology},
  author={Boer, Almut and Nischal, KC},
  journal={Indian Journal of Dermatology, Venereology and Leprology},
  volume={73},
  pages={138},
  year={2007},
  publisher={scientific scholar}
}

@article{giotis2015med,
  title={MED-NODE: A computer-assisted melanoma diagnosis system using non-dermoscopic images},
  author={Giotis, Ioannis and Molders, Nynke and Land, Sander and Biehl, Michael and Jonkman, Marcel F and Petkov, Nicolai},
  journal={Expert systems with applications},
  volume={42},
  number={19},
  pages={6578--6585},
  year={2015},
  publisher={Elsevier}
}

@misc{argenziano2002dermoscopy,
  title={Dermoscopy: a tutorial. EDRA},
  author={Argenziano, G and Soyer, H and De Giorgi, V and Piccolo, D and Carli, P and Delfino, M and others},
  year={2002},
  publisher={Medical Publishing \& New Media Milan, Italy:}
}

@article{rotemberg2021patient,
  title={A patient-centric dataset of images and metadata for identifying melanomas using clinical context},
  author={Rotemberg, Veronica and Kurtansky, Nicholas and Betz-Stablein, Brigid and Caffery, Liam and Chousakos, Emmanouil and Codella, Noel and Combalia, Marc and Dusza, Stephen and Guitera, Pascale and Gutman, David and others},
  journal={Scientific data},
  volume={8},
  number={1},
  pages={34},
  year={2021},
  publisher={Nature Publishing Group UK London}
}

@article{combalia2019bcn20000,
  title={Bcn20000: Dermoscopic lesions in the wild},
  author={Combalia, Marc and Codella, Noel CF and Rotemberg, Veronica and Helba, Brian and Vilaplana, Veronica and Reiter, Ofer and Carrera, Cristina and Barreiro, Alicia and Halpern, Allan C and Puig, Susana and others},
  journal={arXiv preprint arXiv:1908.02288},
  year={2019}
}

@article{tschandl2018ham10000,
  title={The HAM10000 dataset, a large collection of multi-source dermatoscopic images of common pigmented skin lesions},
  author={Tschandl, Philipp and Rosendahl, Cliff and Kittler, Harald},
  journal={Scientific data},
  volume={5},
  number={1},
  pages={1--9},
  year={2018},
  publisher={Nature Publishing Group}
}

@article{codella2019skin,
  title={Skin lesion analysis toward melanoma detection 2018: A challenge hosted by the international skin imaging collaboration (isic)},
  author={Codella, Noel and Rotemberg, Veronica and Tschandl, Philipp and Celebi, M Emre and Dusza, Stephen and Gutman, David and Helba, Brian and Kalloo, Aadi and Liopyris, Konstantinos and Marchetti, Michael and others},
  journal={arXiv preprint arXiv:1902.03368},
  year={2019}
}

@inproceedings{codella2018skin,
  title={Skin lesion analysis toward melanoma detection: A challenge at the 2017 international symposium on biomedical imaging (isbi), hosted by the international skin imaging collaboration (isic)},
  author={Codella, Noel CF and Gutman, David and Celebi, M Emre and Helba, Brian and Marchetti, Michael A and Dusza, Stephen W and Kalloo, Aadi and Liopyris, Konstantinos and Mishra, Nabin and Kittler, Harald and others},
  booktitle={2018 IEEE 15th international symposium on biomedical imaging (ISBI 2018)},
  pages={168--172},
  year={2018},
  organization={IEEE}
}

@article{gutman2016skin,
  title={Skin lesion analysis toward melanoma detection: A challenge at the international symposium on biomedical imaging (ISBI) 2016, hosted by the international skin imaging collaboration (ISIC)},
  author={Gutman, David and Codella, Noel CF and Celebi, Emre and Helba, Brian and Marchetti, Michael and Mishra, Nabin and Halpern, Allan},
  journal={arXiv preprint arXiv:1605.01397},
  year={2016}
}

@article{ballerini2013color,
  title={A color and texture based hierarchical K-NN approach to the classification of non-melanoma skin lesions},
  author={Ballerini, Lucia and Fisher, Robert B and Aldridge, Ben and Rees, Jonathan},
  journal={Color medical image analysis},
  pages={63--86},
  year={2013},
  publisher={Springer}
}

@inproceedings{mendoncca2013ph,
  title={PH 2-A dermoscopic image database for research and benchmarking},
  author={Mendon{\c{c}}a, Teresa and Ferreira, Pedro M and Marques, Jorge S and Marcal, Andr{\'e} RS and Rozeira, Jorge},
  booktitle={2013 35th annual international conference of the IEEE engineering in medicine and biology society (EMBC)},
  pages={5437--5440},
  year={2013},
  organization={IEEE}
}

@article{karimkhani2017global,
  title={Global skin disease morbidity and mortality: an update from the global burden of disease study 2013},
  author={Karimkhani, Chante and Dellavalle, Robert P and Coffeng, Luc E and Flohr, Carsten and Hay, Roderick J and Langan, Sin{\'e}ad M and Nsoesie, Elaine O and Ferrari, Alize J and Erskine, Holly E and Silverberg, Jonathan I and others},
  journal={JAMA dermatology},
  volume={153},
  number={5},
  pages={406--412},
  year={2017},
  publisher={American Medical Association}
}

@misc{flohr2021putting,
  title={Putting the burden of skin diseases on the global map},
  author={Flohr, C and Hay, RJBJoD},
  journal={British Journal of Dermatology},
  volume={184},
  number={2},
  pages={189--190},
  year={2021},
  publisher={Blackwell Publishing Ltd Oxford, UK}
}

@article{zhou2024pre,
  title={Pre-trained multimodal large language model enhances dermatological diagnosis using SkinGPT-4},
  author={Zhou, Juexiao and He, Xiaonan and Sun, Liyuan and Xu, Jiannan and Chen, Xiuying and Chu, Yuetan and Zhou, Longxi and Liao, Xingyu and Zhang, Bin and Afvari, Shawn and others},
  journal={Nature Communications},
  volume={15},
  number={1},
  pages={5649},
  year={2024},
  publisher={Nature Publishing Group UK London}
}

@article{choy2023systematic,
  title={Systematic review of deep learning image analyses for the diagnosis and monitoring of skin disease},
  author={Choy, Shern Ping and Kim, Byung Jin and Paolino, Alexandra and Tan, Wei Ren and Lim, Sarah Man Lin and Seo, Jessica and Tan, Sze Ping and Francis, Luc and Tsakok, Teresa and Simpson, Michael and others},
  journal={NPJ Digital Medicine},
  volume={6},
  number={1},
  pages={180},
  year={2023},
  publisher={Nature Publishing Group UK London}
}

@article{thieme2023deep,
  title={A deep-learning algorithm to classify skin lesions from mpox virus infection},
  author={Thieme, Alexander H and Zheng, Yuanning and Machiraju, Gautam and Sadee, Chris and Mittermaier, Mirja and Gertler, Maximilian and Salinas, Jorge L and Srinivasan, Krithika and Gyawali, Prashnna and Carrillo-Perez, Francisco and others},
  journal={Nature medicine},
  volume={29},
  number={3},
  pages={738--747},
  year={2023},
  publisher={Nature Publishing Group US New York}
}

@article{jaech2024openai,
  title={Openai o1 system card},
  author={Jaech, Aaron and Kalai, Adam and Lerer, Adam and Richardson, Adam and El-Kishky, Ahmed and Low, Aiden and Helyar, Alec and Madry, Aleksander and Beutel, Alex and Carney, Alex and others},
  journal={arXiv preprint arXiv:2412.16720},
  year={2024}
}

@article{zhang2023multimodal,
  title={Multimodal chain-of-thought reasoning in language models},
  author={Zhang, Zhuosheng and Zhang, Aston and Li, Mu and Zhao, Hai and Karypis, George and Smola, Alex},
  journal={arXiv preprint arXiv:2302.00923},
  year={2023}
}

@article{liu2024medcot,
  title={Medcot: Medical chain of thought via hierarchical expert},
  author={Liu, Jiaxiang and Wang, Yuan and Du, Jiawei and Zhou, Joey Tianyi and Liu, Zuozhu},
  journal={arXiv preprint arXiv:2412.13736},
  year={2024}
}

@article{yan2025multimodal,
  title={A multimodal vision foundation model for clinical dermatology},
  author={Yan, Siyuan and Yu, Zhen and Primiero, Clare and Vico-Alonso, Cristina and Wang, Zhonghua and Yang, Litao and Tschandl, Philipp and Hu, Ming and Ju, Lie and Tan, Gin and others},
  journal={Nature Medicine},
  pages={1--12},
  year={2025},
  publisher={Nature Publishing Group}
}
}

\end{document}